\documentclass[letterpaper, 10 pt, conference]{ieeeconf}  

\IEEEoverridecommandlockouts                              
                                                          
\usepackage{graphicx} 
\usepackage{amsmath} 
\usepackage{amssymb}  
\usepackage{caption}
\usepackage{subcaption}
\usepackage{tikz}
\newcommand{\Cov}[1]{\textnormal{Cov} \left[ #1 \right]}
\newcommand{\prob}[1]{\textnormal{p} \left( #1 \right)}

\title{\LARGE \bf Learning Rank Reduced Interpolation \\ with Principal Component Analysis}
\author{Matthias Ochs$^{1}$, Henry Bradler$^{1}$ and Rudolf Mester$^{1,2}$
\thanks{$^{1}$Visual Sensorics \& Information Processing Lab, Goethe University, Frankfurt am Main, Germany}%
\thanks{$^{2}$Computer Vision Laboratory, ISY, Link\"oping University, Sweden}%
}

\begin{document}

\maketitle

\begin{tikzpicture}[remember picture,overlay]
\node[anchor=south,yshift=10pt] at (current page.south) {\fbox{\parbox{\dimexpr\textwidth-\fboxsep-\fboxrule\relax}{\footnotesize \textcopyright 2017 IEEE. Personal use of this material is permitted.
			Permission from IEEE must be obtained for all other uses, in any current or future
			media, including reprinting/republishing this material for advertising or promotional
			purposes, creating new collective works, for resale or redistribution to servers or
			lists, or reuse of any copyrighted component of this work in other works.
			DOI: pending}}};
\end{tikzpicture}

\thispagestyle{empty}
\pagestyle{empty}

\begin{abstract}
	
In computer vision most iterative optimization algorithms, both sparse and dense,
rely on a coarse and reliable dense initialization to bootstrap their optimization procedure.
For example, dense optical flow algorithms profit massively
in speed and robustness
if they are initialized well in the basin of convergence
of the used loss function.
The same holds true for methods as sparse feature tracking 
when initial flow or depth information for new features at arbitrary positions is needed.
This makes it extremely important to have techniques at hand that 
allow to obtain 
from only very few available measurements
a dense but still approximative 'sketch' of
a desired 2D structure
(e.g.\ depth maps, optical flow, disparity maps, etc.).

The 2D map
is regarded as sample from a 2D random process.
The  method presented here exploits the complete information
given by the principal component analysis (PCA) of that process,
the principal basis \emph{and} its prior distribution.
The method is able to determine a dense reconstruction from sparse measurement.
When facing situations with only very sparse measurements, 
typically the number of principal components 
is further reduced which results in a loss of expressiveness of the basis.
We overcome this problem and inject prior knowledge in a maximum a posterior (MAP) approach.

We test our approach on the KITTI and the virtual KITTI datasets
and focus on the interpolation of depth maps for driving scenes.
The evaluation of the results show good agreement to the ground truth
and are clearly better than results of interpolation by the nearest neighbor method
which disregards statistical information.
	
\end{abstract}
\section{Introduction}

This paper addresses the task of computing a dense reconstruction 
of depth images as they are typically of interest
in visual odometry and SLAM. This task is challenging
if only a very sparse measurement of the depth is given.
We focus on the situation where
depth (or disparity) information is available for less than 1/1000 of all pixels (e.g.\ for about $200$ pixels for an image of size $1240 \times 370$).
As we will see, it is possible
to tackle this problem sucessfully if a suitable statistical
model exists for the regarded class of depth images.


It may seem surprising that this problem can be addressed
using a Bayesian version of PCA. 
Determining a PCA model offers \emph{two} powerful approaches:
1) reducing the rank of the model (subspace model)
2) using the eigenvalues of the initial processes covariance matrix
	reflected in the variances of the PCA model.

In this paper we demonstrate how to overcome this indeterminacy by including prior knowledge about the principal components which is already given by the PCA.
We determine a best set of coefficients for a representation in the principal coordinate system by utilizing a maximum a posteriori estimation (MAP).
Furthermore,
we demonstrate how this approach can be used to get a dense reconstruction of data for which only very sparse information is available and which though preserves the coarse structure and the main print of the dense original data.

The structure of this paper is as follows.
In the approach part we show how to learn the basis of principal components for a specific class of data.
In our case we  deal with depth maps for an application in an automotive scenario,
but of course the method is not limited to this exemplary case.
For a sparse measurement,
we then derive the best representation in a given PCA basis from a statistical approach which exploits prior knowledge about the general distribution of the principal components.
The interpolation step consists of a transformation from the principal coordinate system back to the system of the original data.
This finally yields the dense reconstruction sought.

\begin{figure}[t!]
	\centering
	\includegraphics[width=1.\linewidth]{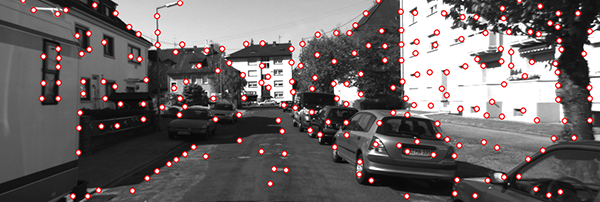} \\
	\vspace{0.1cm}
	\includegraphics[width=1.\linewidth]{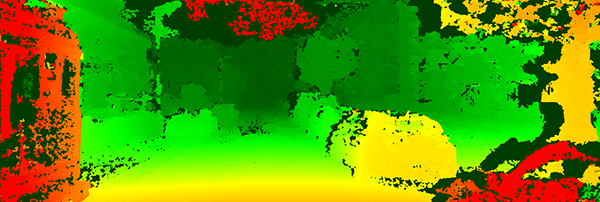} \\
	\vspace{0.1cm}
	\includegraphics[width=1.\linewidth]{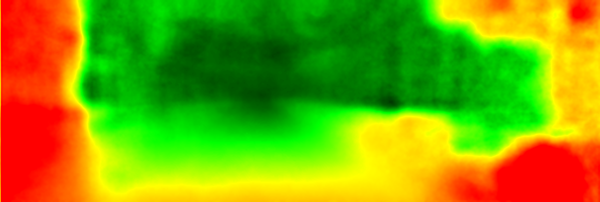} 
	\caption{The top image shows frame 532 of the KITTI test sequence 19 and indicates where sparse measurements of the depth have been obtained (circles).
    The middle image shows, as reference, a SGBM depth map.
    Our interpolation of the sparse measurements is presented at the bottom.}
	\label{fig:intro:teaser}
\end{figure}

In the experimental part, we  evaluate our approach on two familiar datasets. 
We regards the well known KITTI odometry dataset
 \cite{GeigerLenzUrtasun2012CVPR} 
and the more recent synthetic \emph{Virtual KITTI} dataset \cite{GaidonWangCabonVig2016}.
The latter has the benefit of providing pixelwise ground truth of the depth structure.
We investigate also the applicability of our approach by comparing the reconstructed 3D point cloud and pixel correspondences (stereo for KITTI / temporal for Virtual KITTI) against reference values (SGBM for KITTI / groundtruth for Virtual KITTI).
The pixelwise evaluation of our dense interpolation shows very good results and that it is clearly superior to interpolations which do not make use of statistical approaches as, e.g., nearest neighbour interpolation.
An example of dense reconstruction obtained by our method is shown in figure \ref{fig:intro:teaser}.
\section{Related Work}

Monocular visual odometry / SLAM algorithms 
became more and more popular recently
for perception of the environment in autonomous driving.
All those algorithms have in common that their performance
is strongly dependent on a good initialization of the relative pose and feature points between two consecutive frames.
Engel et al.\ \cite{EngelSchoepsCremers2014} proposed in their LSD-SLAM approach a randomly generated depth map,
 which is improved with valid measurements and propagated during tracking with key frames.
A similar approach is pursued by Forster et al.\ \cite{ForsterPizzoliScaramuzza2014}. 
They maintain the depth maps for initialization with Bayesian depth filters and matching desired feature points.
The authors of the propagation based tracking method \cite{FananiOchsBradlerMester2016IV}
initialize new feature points based on the displacements computed by phase correlation.
The depth values of feature points that have been tracked at least once before
are propagated into the next frame, using epipolar matching,
an initial estimate of the next relative pose and a subsequent
joint optimization of this pose and the point correspondences
\cite{BradlerOchsMester2017WACV}.

All these monocular SLAM methods have in common that they only reconstruct a sparse set of points.
For autonomous driving, it is obviously not sufficient to have
3D information only at some few points --- densification is necessary.
In order to allow a densification process to converge quickly
into a true detailed dense map, it is useful to
have a solid method that generates initial estimates
of these dense depth maps.

Also for many dense optical flow methods \cite{BroxBruhnPapenbergWeickert2004,Horn81Schunck1981},
it is advantageous to have a good initialization for the dense depth map,
since computationally expensive variational approaches
converge faster if they are initialized close to the optimum.

Since estimating a depth map or an optical flow field 
is almost the same for rigid scenes and a given relative pose,
we compare our method also to related work in the field of optical flow estimation.

There are dense optical flow methods which also use sparse measurements for initialization.
For example, Gibson and Span \cite{GibsonSpann2003} uses a sparse feature tracking algorithm in their first stage.
In their second step, a traditional dense optical flow optimization is followed.
SIFT Flow \cite{LiuYuenTorralba2011} uses densely sampled SIFT features, which are matched between two images.
These matches are then used to compute a dense optical flow field.
Leordeanu et al.\ \cite{LeordeanuZanfirSminchisescu2013} also used sparse measurements
to initialize their dense optical flow method.
They use a sparse set of correspondences to perform a sparse-to-dense interpolation
which are then refined using a total variation model.
Wulff and Black \cite{WulffBlack2015} also calculate dense optical flow
given a set of sparse feature point matches.
These points are used to estimate several PCA flow field layers.
The combination of these layers into a dense optical flow field is done with an MRF.
This approach is very similar to our PCA interpolation of depth maps.
In contrast to them, we use a maximum a-posteriori estimation
to estimate weighted linear combination of the PCA basis.
This leads to significantly better interpolated depth maps.

Robert et at. \cite{RobertsPotthastDellaert2009} estimate
 the optical flow field and the ego-motion based on a probabilistic PCA.
This approach is extended in the work by Herdtweck and Curio \cite{HerdtweckCurio2012} into the so-called 'expert models'.
Each of those export models represents a specific pre-trained subspace of the training data.
The optical flow field and ego-motion are estimated by an expert system and an outlier model.
Another learning-based method for estimating a depth map are proposed by Saxena et al.\ \cite{SaxenaChungNg2005}.
They learn discriminatively a Markov Random Field at multiple spatial scales to predict depth maps.

Besides all those 'classical' computer vision approaches,
there also exists methods that compute depth maps or optical flow fields with deep learning networks.
Eigen et al.\ \cite{EigenPuhrschFergus2014} presented one of the first works which use deep learning to estimate depth maps.
This approach was improved and extended to compute surface normals and also semantic labels in \cite{EigenFergus2015}.
All these deep learning techniques have the disadvantage that they need a huge amount of precise ground truth data for training.
Garg et al. \cite{GargKumarCarneiroReid2016} proposed an unsupervised framework
to train a CNN for estimation of a depth map from a single image without the need of annotated ground truth depths. 
Another drawback of the deep learning networks
has recently been solved by Mancini et al.\ \cite{ManciniCostanteValigiCiarfugliaDelmericaScaramuzza2017}.
They included a \emph{long short term memory} into their network.
By this, it is possible to estimate the global scale of the depth maps.
Previously, this was a problem since a deep network was only capable to estimate a unscaled depth map from a monocular images.

\section{Approach}

Our densification / interpolation scheme consists of the following three steps:
\begin{itemize}
	\item learning a basis (once for a specific class of data)
	\item projection to this basis (once for each sparse measurement)
	\item interpolation by projection back to original basis (once for each sparse measurement)
\end{itemize}
The first step only has to be performed once for each class of data considered.
Thus, we need to learn one basis which is suitable to represent depth maps
as they are typically observed in an automotive driving scenario \cite{GeigerLenzUrtasun2012CVPR, Cordts2016Cityscapes}.
This learning step can be done offline, utilizing a large training data set (like the depth data of the complete KITTI dataset) that represents the typical statistics of the data to be expected in use.

Given this precomputed  basis,
any set of sparse measurement data
needs to be associated with the coefficients of the basis
such that the linear superposition of the basis vector (=basis images)
complies with the sparse measured data,
either perfectly, or as good as possible in the sense
of a suitable metric.
In our scenario, the number of basis vectors will, in general, by far exceed the number of measurements (underdetermined problem).
This is not a problem as long as we include prior knowledge about the coefficient distribution.
We show that this prior knowledge is already available from the learning step.
This is similiar to the approach of \cite{WulffBlack2015}.

Finally, the representation of the sparse measurement in the learnt basis is used to transform the coefficients back to the original domain. This gets a dense interpolation.

\textbf{Note:} When using the word \emph{depth} in this paper,
we generally mean a quantity which is suitable to express the depth.
E.g., this also applies for the inverse depth or disparity
which both possesses advantages over the actual depth
when it comes to statistical properties and numerical representation.
If depth is encoded as disparity, the conversion factor (depends on focal length and base of a stereo setup) to inverse depth must be given.
Even though disparity in general is used in stereo scenarios,
it can be numerically beneficial to store depth or inverse depth as disparity in a non stereo setup
by choosing a reference stereo base for conversion.

\subsection{Learning a Basis via PCA}
\label{sec:basis-learning}

We start with a training set of $n$ depth maps $\vec{d}_{i} \in \mathbb{R}^{s}$,
each of dimension $s$, which cover the typical range variation which could be expected in the scenario.
In our case of high-resolution depth maps the number of training samples is much smaller than their dimension $n \ll s$.
Thus, at most $n$ of the potential $s$ degrees of freedom of the data can be revealed.
However, for images of a specific class (e.g.\ depth maps),
in contrast to random permutations,
already a few principal components are sufficient to express the coarse impression and the main information of the image.

To find these significant degrees of freedom, we employ PCA on the training set.
For that purpose, the mean and covariance need to be computed.
The unbiased sample mean and covariance of the training set are given by:
\begin{align}
	\label{eq:sample-mean}
	\vec{m} &= \frac{1}{n} \sum_{i = 0}^{n - 1} \vec{d_{i}}, \\
	\label{eq:sample-cov}
	\mathbf{C} &= \frac{1}{n - 1} \mathbf{D} \cdot \mathbf{D}^{T} \in \mathbb{R}^{s \times s} \quad \textnormal{, with} \\
	\mathbf{D} &= \left( \vec{d}_{0} - \vec{m}, \ldots, \vec{d}_{n - 1} - \vec{m} \right) \in \mathbb{R}^{s \times n}.
\end{align}
The spectral decomposition of the covariance yields
\begin{align}
	\label{eq:spectral-decomposition}
	\mathbf{C} &= \mathbf{U} \cdot \textnormal{diag}\left( \lambda_{0}, \ldots, \lambda_{s - 1} \right) \cdot \mathbf{U}^{T}, \\
	\mathbf{U} &= \left( \vec{u}_{0}, \ldots, \vec{u}_{s - 1} \right) \in \mathbb{R}^{s \times s},
\end{align}
with the principal components and the corresponding variances stored as basis vectors $\vec{u}_{i}$ in the columns of $\mathbf{U}$ and as eigenvalues in descending order in the diagonal matrix $\textnormal{diag}(\lambda_{0}, \ldots, \lambda_{s - 1})$, $\lambda_{0} \ge \ldots \ge \lambda_{s - 1}$, respectively.
So, for a given depth map $\vec{d}$ the basis coefficients $\vec{y}$ of the principal coordinate system are
\begin{align}
	\label{eq:principal-coordinate-system}
	\vec{y} = \mathbf{U}^{T} \cdot \left( \vec{d} - \vec{m} \right) \quad , \quad \vec{d} = \mathbf{U} \cdot \vec{y} + \vec{m}.
\end{align}
From equation \eqref{eq:sample-cov} it becomes obvious that $\mathbf{C}$ is rank deficient,
$\textnormal{rank}(\mathbf{C}) \le n \ll s$,
and since the sample covariance is computed using the sample mean we even know $\textnormal{rank}(\mathbf{C}) < n$ or rather $\lambda_{i \ge (n - 1)} = 0$.
This implies that the PCA of the training set allows to uncover at most $n - 1$ degrees of freedom,
 the mean and $n - 2$ principal components which encode information (nonzero variance).

\begin{figure}[t!]
	\centering
	\begin{subfigure}[b]{0.49\linewidth}
		\includegraphics[width=\linewidth]{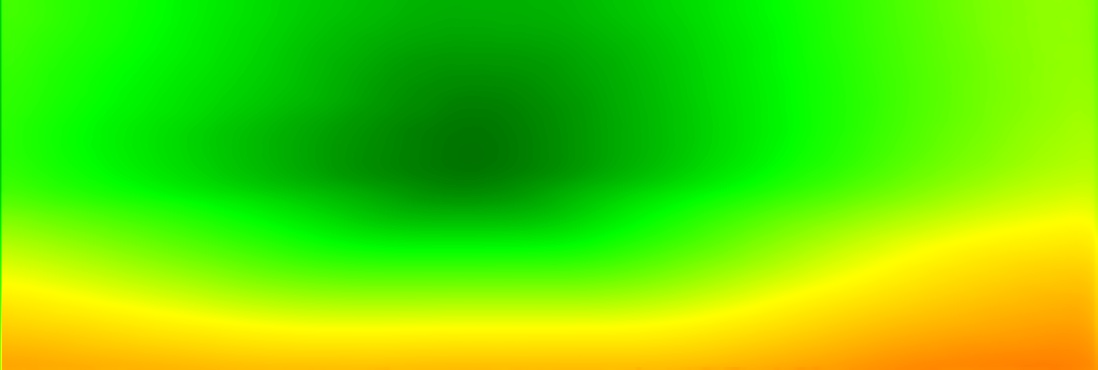}
		\caption{$\vec{m}$}
		\label{fig:pca-mean}
	\end{subfigure}
	\vspace{0.1cm}
	\begin{subfigure}[b]{0.49\linewidth}
		\includegraphics[width=\linewidth]{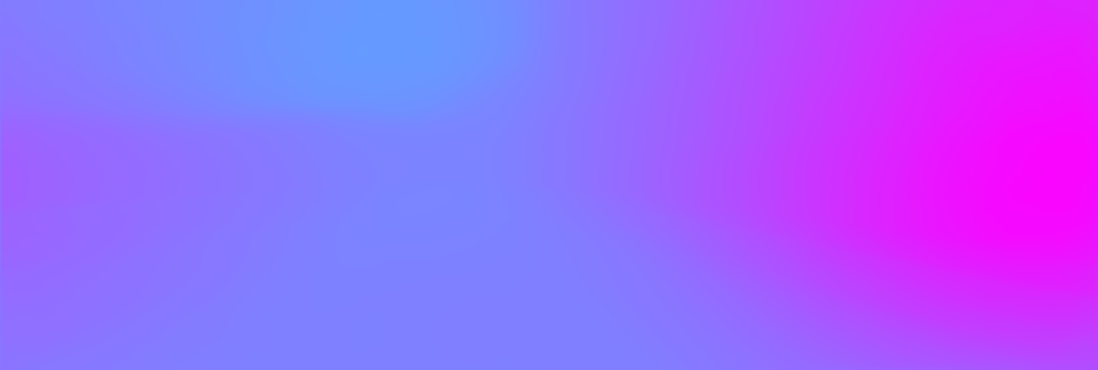}
		\caption{$\vec{u}_{0}$}
		\label{fig:pca-eigen0}
	\end{subfigure} \\
	\begin{subfigure}[b]{0.49\linewidth}
		\includegraphics[width=\linewidth]{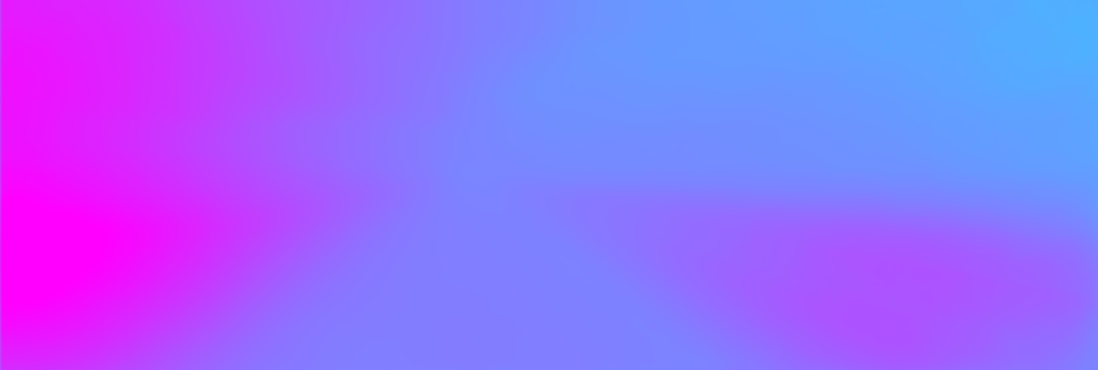}
		\caption{$\vec{u}_{1}$}
		\label{fig:pca-eigen1}
	\end{subfigure}
	\vspace{0.1cm}
	\begin{subfigure}[b]{0.49\linewidth}
		\includegraphics[width=\linewidth]{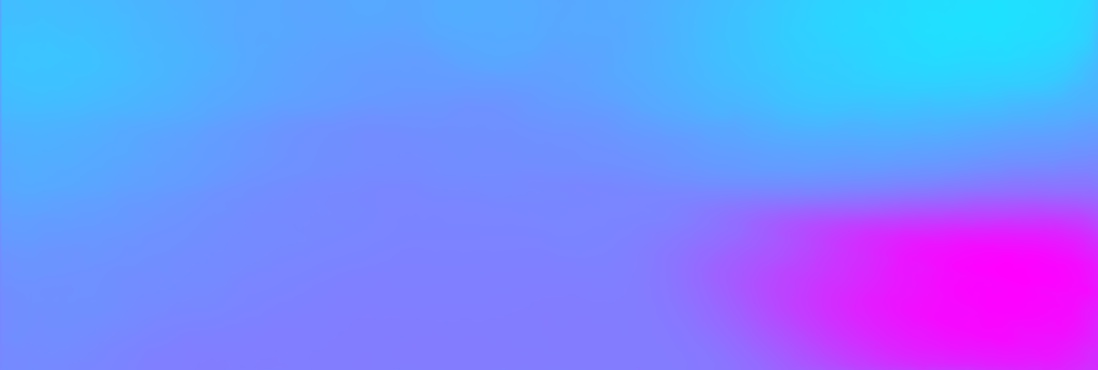}
		\caption{$\vec{u}_{2}$}
		\label{fig:pca-eigen2}
	\end{subfigure}
	\caption{Mean depth and the three most significant basis vectors computed via the PCA of the depth maps of all KITTI training sequences (00-10).}
	\label{fig:approach:pca-basis}
\end{figure}

\begin{figure}[ht]
	\centering
	\includegraphics[width=.9\linewidth]{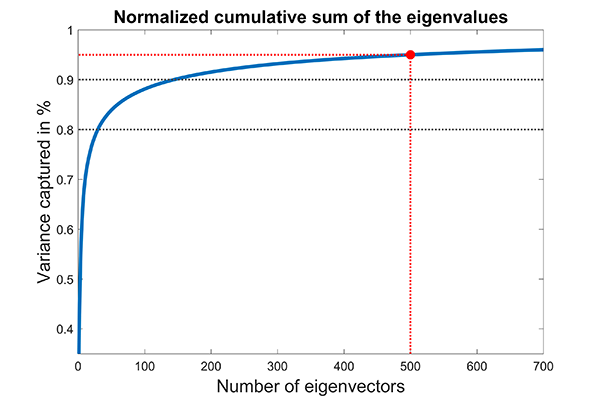} \\
	\caption{The relative cumulative sum shows
	how much variance of the learnt depth map can be reconstructed
	by using a certain number of the largest eigenvalues and their corresponding eigenvectors.}
	\label{fig:experiments:cumsum}
\end{figure}

However, for our aim of a coarse but dense interpolation,
we do not need a complete basis of principal components.
If we examine the cumulative sum of the variances $\lambda_{i}$ (see figure \ref{fig:experiments:cumsum}),
a reasonable limit $l$ of basis vectors to consider can be determined.
We restrict ourselves to a basis that consists of the $l \ll n \ll s$ most important basis vectors that are capable of explaining more than $90\%$ of the information (variance) of the training set.
All depth maps $\vec{d}$ of the same class as the training samples possess a sufficiently good approximation in this \emph{truncated basis}.
\begin{align}
	\mathbf{B} &= \left( \vec{u}_{0}, \ldots, \vec{u}_{l - 1} \right), \\
	\mathbf{\Lambda} &= \textnormal{diag}\left( \lambda_{0}, \ldots, \lambda_{l - 1} \right).
\end{align}
Figure \ref{fig:approach:pca-basis} shows an example of the mean depth and the most significant principal components.

\subsection{Projection to the Principal Coordinate System}
\label{sec:projection-pcs}

How do we find the coefficients $\vec{y}$ that are most suited to express a sparse measurement $\tilde{\vec{z}} = \tilde{\vec{d}} - \tilde{\vec{m}}$ (the tilde denotes that only some few components of the vector are given)?

In our application, we face a situation where the number of measured components of the depth map
 is even below the size $l$ of our reduced basis $\mathbf{B}$.
In this case a simple (weighted) least squares approach to compute the coefficient vector $\vec{y}$ via the minimization of the difference of measured and reconstructed signal would lead to an under-determined problem.
We take care of this by including prior knowledge about the coefficients which is already given by the eigenvalues $\mathbf{\Lambda}$ of the sample covariance $\mathbf{C}$.

This leads to a maximum a posteriori (MAP) estimation of $\vec{y}$,
where the maximization of the posterior,
\begin{align}
	\label{eq:map}
	\prob{\vec{y} \,|\, \tilde{\vec{z}}} &= \alpha \cdot \prob{\tilde{\vec{z}} \,|\, \vec{y}} \cdot \prob{\vec{y}} \rightarrow \max,
\end{align}
is equivalent to the minimizing of its negative logarithm
\begin{align}
	\nonumber
	-\log\left( \prob{\vec{y} \,|\, \tilde{\vec{z}}} \right) &=  \alpha^{'} + \left( \tilde{\vec{z}} - \tilde{\mathbf{B}} \cdot \vec{y} \right)^{T} \cdot \mathbf{C}_{\tilde{\vec{z}}}^{-1} \cdot \left( \tilde{\vec{z}} - \tilde{\mathbf{B}} \cdot \vec{y} \right) \\
	\label{eq:log-map}
	&\phantom{=} + \vec{y}^{T} \cdot \mathbf{C}_{\vec{y}}^{-1} \cdot \vec{y} \rightarrow \min.
\end{align}
The variables $\alpha$ and $\alpha^{'}$ in equation \eqref{eq:map} and \eqref{eq:log-map} collect all terms independent of $\vec{y}$.
The prior covariance is given by the eigenvalues of the sample covariance $\mathbf{C}_{\vec{y}} = \mathbf{\Lambda}$ and the measurement covariance is assumed to be uncorrelated and to possess constant variance for all components of the measurement $\mathbf{C}_{\tilde{\vec{z}}} = \sigma_{\vec{z}}^{2} \cdot \mathbf{I}$.

In terms of the posterior,
the estimate $\hat{\vec{y}}$,
defined by the linear equation system
\begin{align}
		\left(\sigma_{\vec{z}}^{2} \cdot \mathbf{\Lambda}^{-1} + \tilde{\mathbf{B}}^{T} \cdot \tilde{\mathbf{B}} \right) \cdot \hat{\vec{y}} = \tilde{\mathbf{B}}^{T} \cdot \tilde{\vec{z}},		
\end{align}
gives the best representation of the sparse measurement $\tilde{\vec{d}}$ in the PCA coordinate system.
Its covariance is given by
\begin{align}
	\Cov{\vec{y}} = \left(\sigma_{\vec{z}}^{2} \cdot \mathbf{\Lambda}^{-1} + \tilde{\mathbf{B}}^{T} \cdot \tilde{\mathbf{B}} \right)^{-1}.
\end{align}

\subsection{Dense Interpolation}
\label{sec:dense-interpolation}

Once we computed the coefficient vector estimate $\hat{\vec{y}}$ via the MAP approach,
it is straightforward to get the estimate of the reconstructed full depth map $\hat{\vec{d}}$ to the sparse measurement~$\tilde{\vec{d}}$:
\begin{align}
	\hat{\vec{d}} &= \mathbf{B} \cdot \hat{\vec{y}} + \vec{m} \\
	\Cov{\hat{\vec{d}}} &= \mathbf{B} \cdot \Cov{\hat{\vec{y}}} \cdot \mathbf{B}^{T} \\
	\label{eq:full-cov-approx}
	&\approx \sum_{i = 0}^{l - 1} \kappa_{i}^{2} \cdot \vec{u}_{i} \cdot \vec{u}_{i}^{T}.
\end{align}
In equation \eqref{eq:full-cov-approx},
we ignored all possible correlations of $\hat{\vec{y}}$ and only considered the diagonal elements $\kappa_{i}^{2}$ of $\Cov{\hat{\vec{y}}}$ to get an approximation of the covariance of the reconstruction.
Furthermore, we define the uncertainty image $\vec{\xi}$ that consists of the diagonal elements of the approximated covariance as
\begin{align}
	\label{eq:uncertainty}
	\vec{\xi} = \textnormal{diag} \left( \sum_{i = 0}^{l - 1} \kappa_{i}^{2} \cdot \vec{u}_{i} \cdot \vec{u}_{i}^{T} \right).
\end{align}
Apart from the approximations made in equation \eqref{eq:full-cov-approx} and \eqref{eq:uncertainty},
the uncertainty image $\vec{\xi}$ is the propagation of the combined uncertainty of the data (likelihood) and the prior term of the MAP approach to the original image domain of the depth map.
\section{Experiments}

\begin{figure*}[t!]
	\centering
	\includegraphics[width=.24\linewidth]{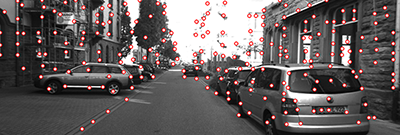} 
	\includegraphics[width=.24\linewidth]{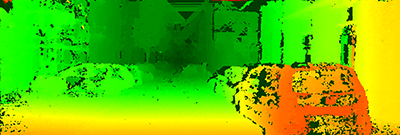}
	\includegraphics[width=.24\linewidth]{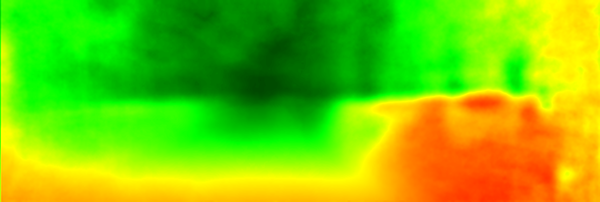}
	\includegraphics[width=.24\linewidth]{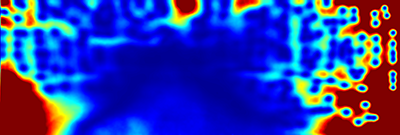} \\
	\vspace{0.1cm}
	\includegraphics[width=.24\linewidth]{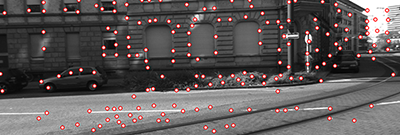} 
	\includegraphics[width=.24\linewidth]{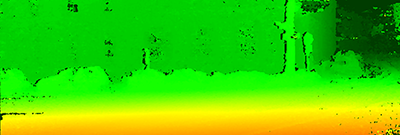}
	\includegraphics[width=.24\linewidth]{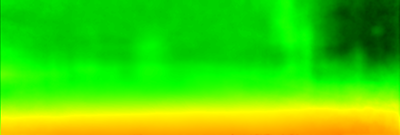}
	\includegraphics[width=.24\linewidth]{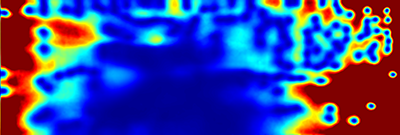} \\
	\vspace{0.1cm}
	\includegraphics[width=.24\linewidth]{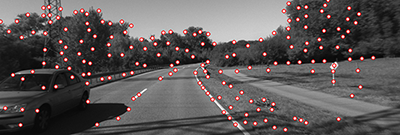} 
	\includegraphics[width=.24\linewidth]{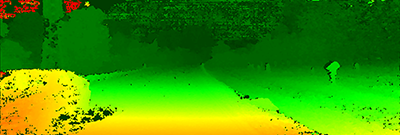}
	\includegraphics[width=.24\linewidth]{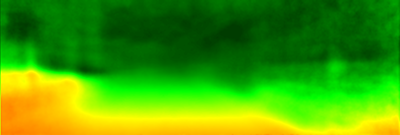}
	\includegraphics[width=.24\linewidth]{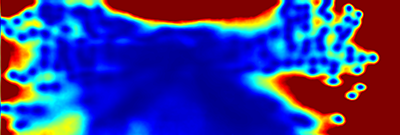} \\
	\vspace{0.1cm}
	\includegraphics[width=.24\linewidth]{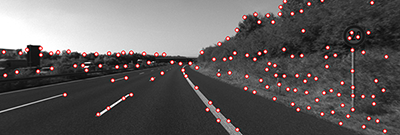} 
	\includegraphics[width=.24\linewidth]{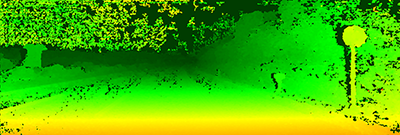}
	\includegraphics[width=.24\linewidth]{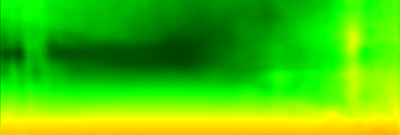}
	\includegraphics[width=.24\linewidth]{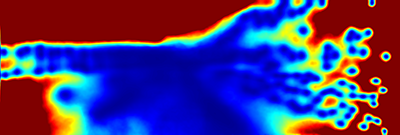}
	\caption{Example results of the KITTI dataset with different driving scenes (urban, rural, highway).
		The first column shows the selected feature points.
		The SGBM depths (second column) are used at these locations 
		to interpolate the depth maps with our approach (third column).
		The uncertainty image of our interpolation
		is shown in the last column (dark red implies high uncertainty).}
	\label{fig:experiments:kitti}
\end{figure*}

\begin{figure*}[t!]
	\centering
	\vspace{0.1cm}
	\includegraphics[width=.24\linewidth]{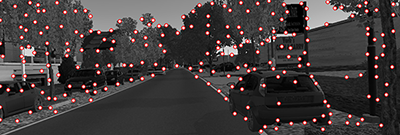} 
	\includegraphics[width=.24\linewidth]{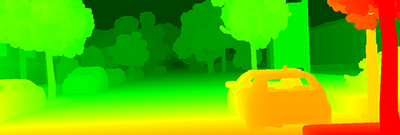}
	\includegraphics[width=.24\linewidth]{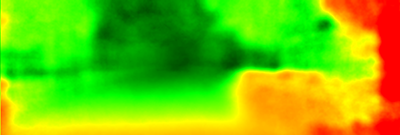}
	\includegraphics[width=.24\linewidth]{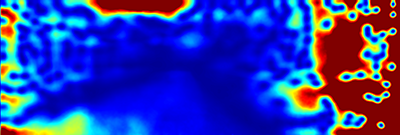} \\
	\vspace{0.1cm}
	\includegraphics[width=.24\linewidth]{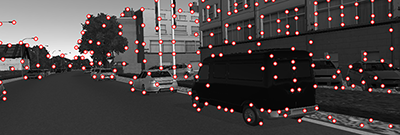} 
	\includegraphics[width=.24\linewidth]{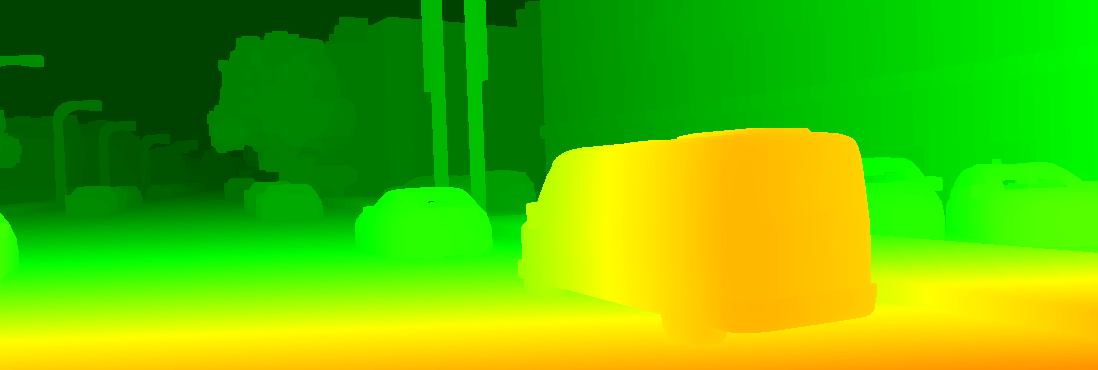}
	\includegraphics[width=.24\linewidth]{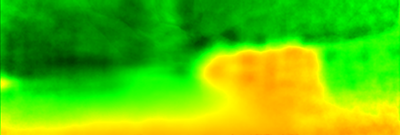}
	\includegraphics[width=.24\linewidth]{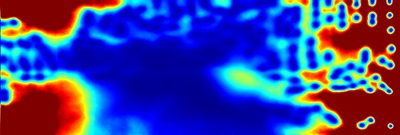} \\
	\vspace{0.1cm}
	\includegraphics[width=.24\linewidth]{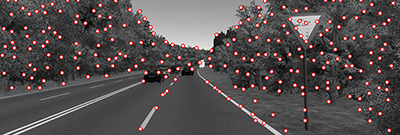} 
	\includegraphics[width=.24\linewidth]{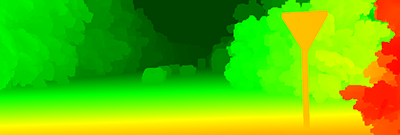}
	\includegraphics[width=.24\linewidth]{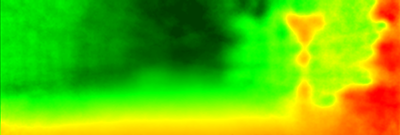}
	\includegraphics[width=.24\linewidth]{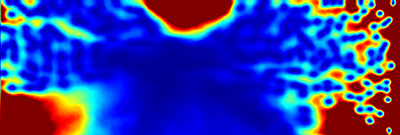} \\
	\vspace{0.1cm}
	\includegraphics[width=.24\linewidth]{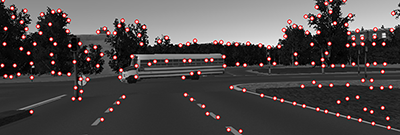} 
	\includegraphics[width=.24\linewidth]{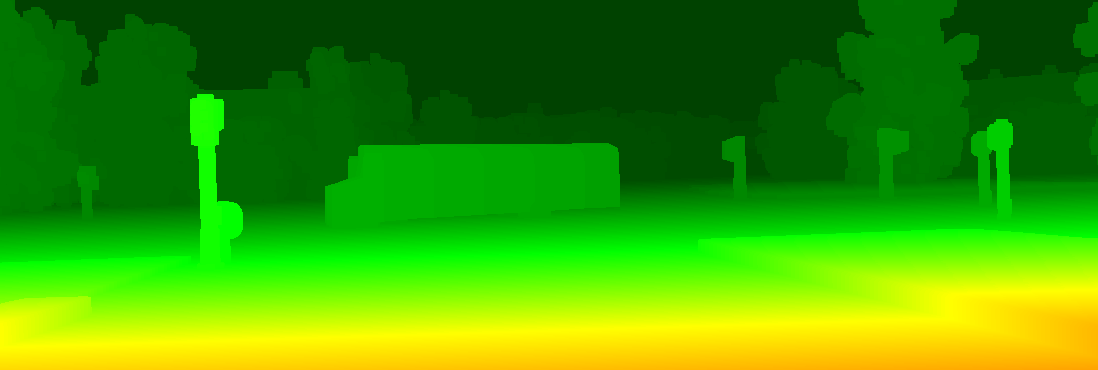}
	\includegraphics[width=.24\linewidth]{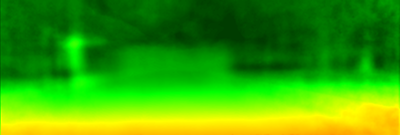}
	\includegraphics[width=.24\linewidth]{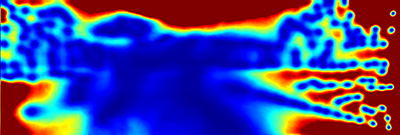}	
	\caption{Example results of the Virtual KITTI dataset with different driving scenes (urban, rural).
		The first column shows the selected feature points. 
		The ground truth depth are presented in the second column. 
		These depth values are used at detected pixel locations to interpolate the depth maps with our approach (third column). 
		The uncertainty image of our interpolation is shown in the last column (dark red implies high uncertainty).}
	\label{fig:experiments:vkitti}
\end{figure*}

\begin{figure*}[t!]
	\centering
	\includegraphics[width=0.32\linewidth]{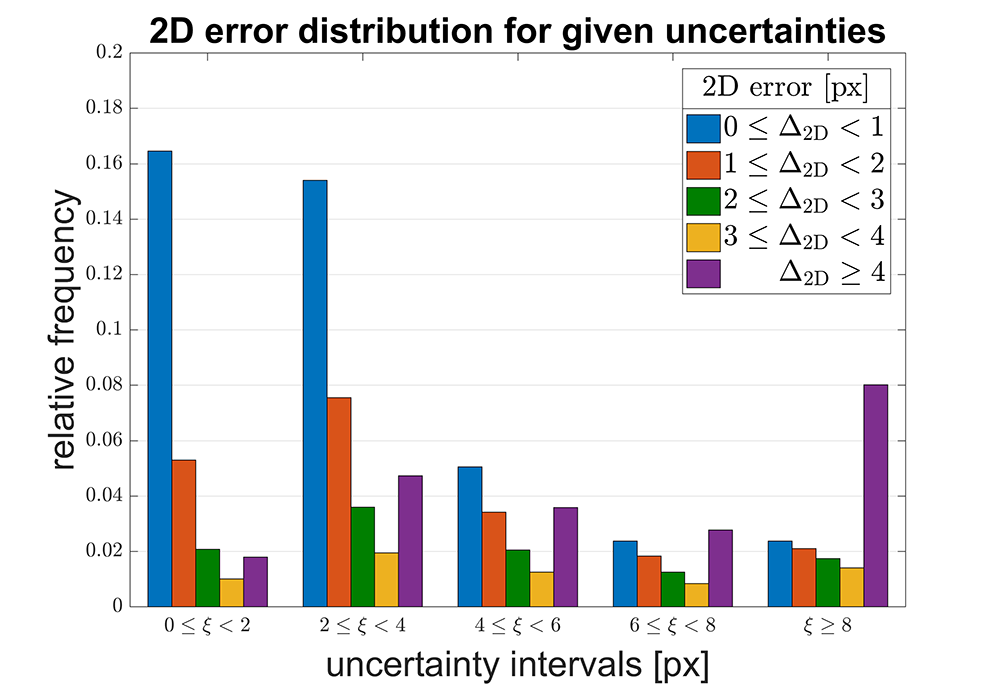}
	\includegraphics[width=0.32\linewidth]{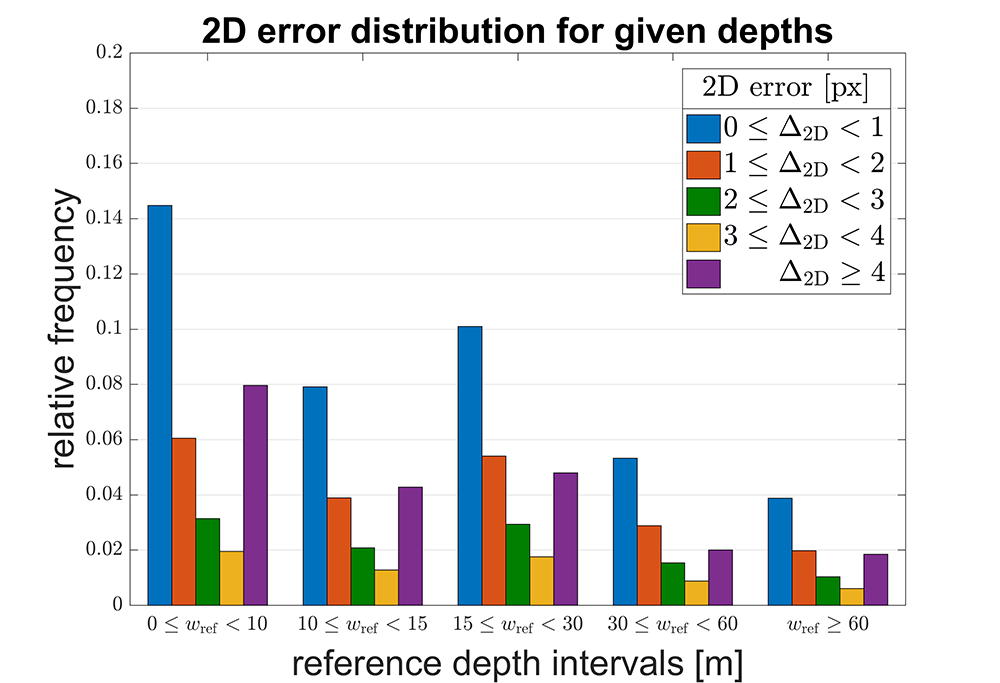}
	\includegraphics[width=0.32\linewidth]{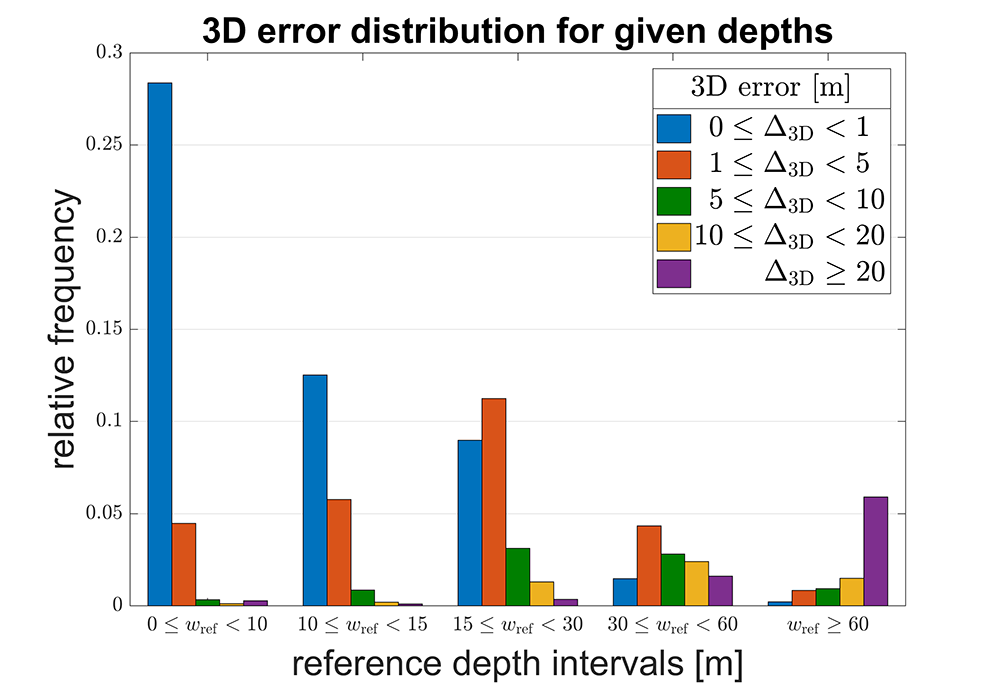} \\
	\vspace{0.1cm}
	\includegraphics[width=0.32\linewidth]{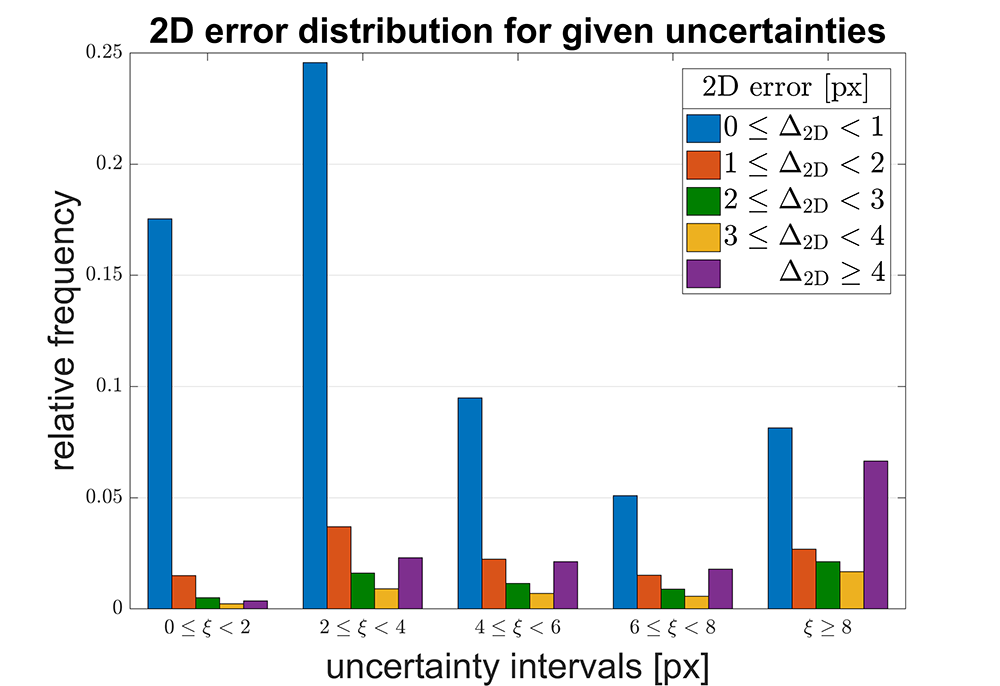}
	\includegraphics[width=0.32\linewidth]{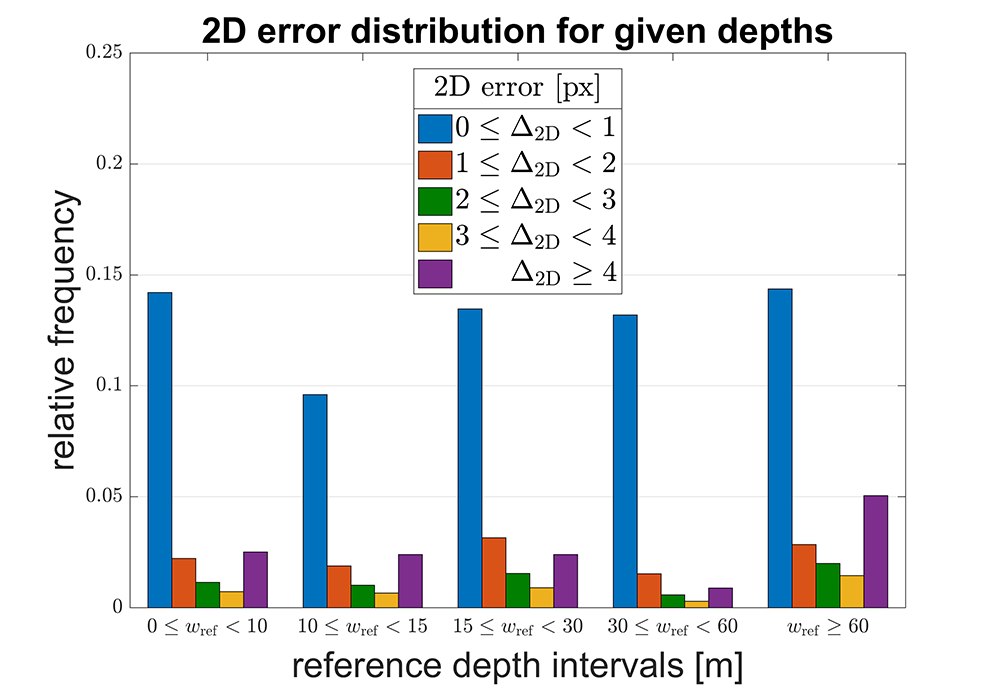}
	\includegraphics[width=0.32\linewidth]{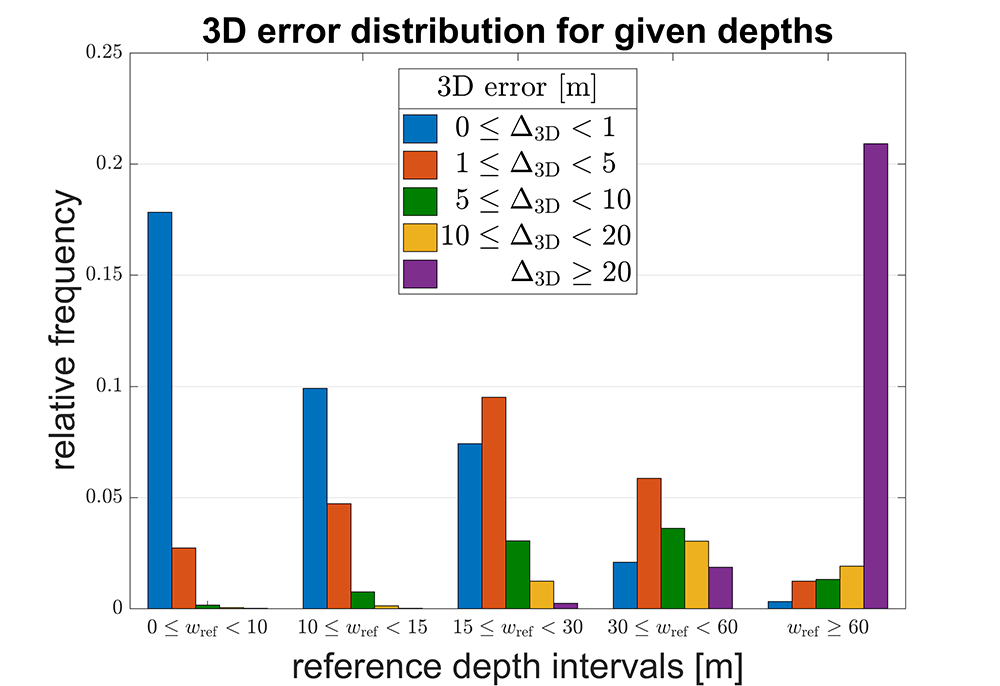}
	\caption{In the first row, the evaluation is performed on the KITTI dataset.
		The second row shows the same evaluations metrics on the Virtual KITTI dataset.
		The 2D error $ \Delta_{2D} $ distribution for different uncertainty levels are shown in the first column.
		In the second and the third column, the 2D error $ \Delta_{2D} $
		respectively the 3D error $ \Delta_{3D} $ exhibits the distributions for different depth bins.}
	\label{fig:experiments:evaluation}
\end{figure*}

\begin{figure*}[ht]
	\centering
	\vspace{0.1cm}
	\includegraphics[width=.24\linewidth]{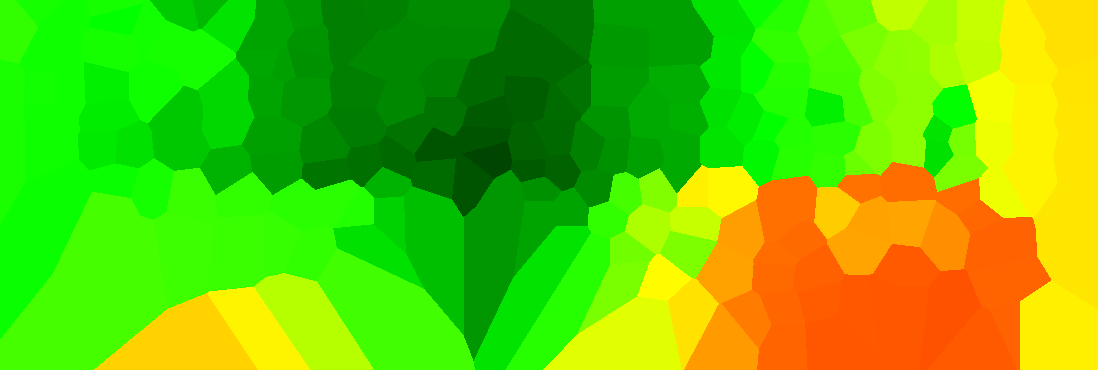} 
	\includegraphics[width=.24\linewidth]{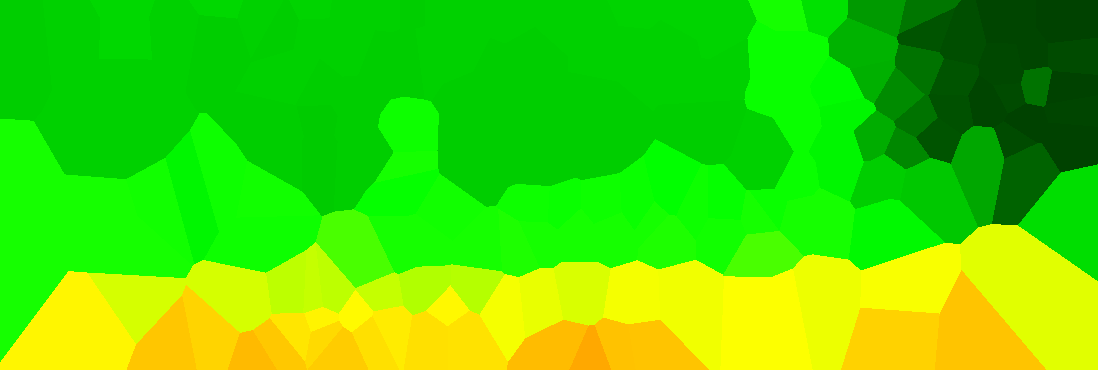}
	\includegraphics[width=.24\linewidth]{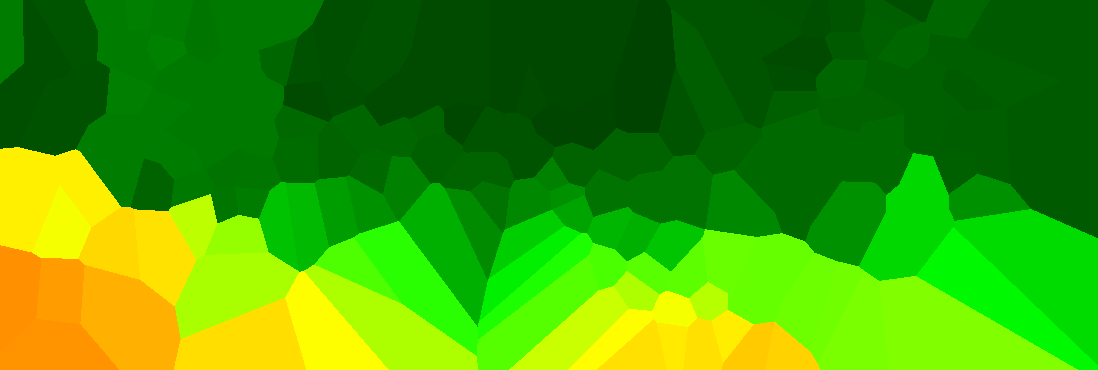}
	\includegraphics[width=.24\linewidth]{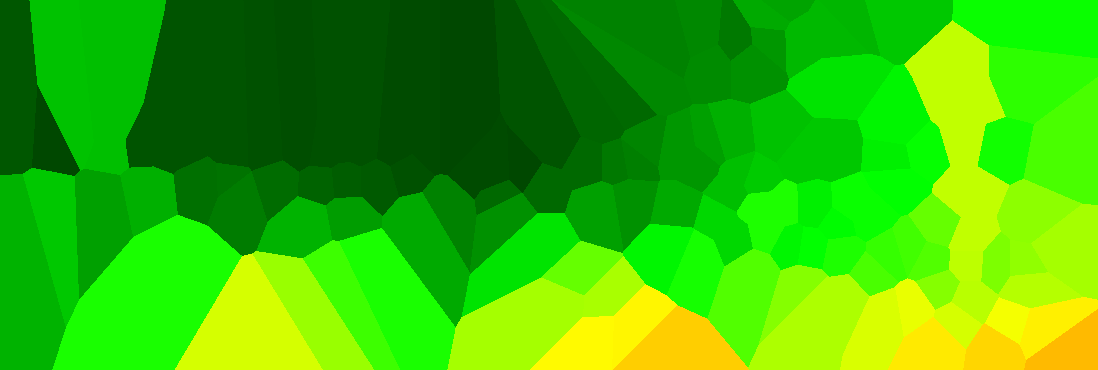} \\
	\vspace{0.1cm}
	\includegraphics[width=.24\linewidth]{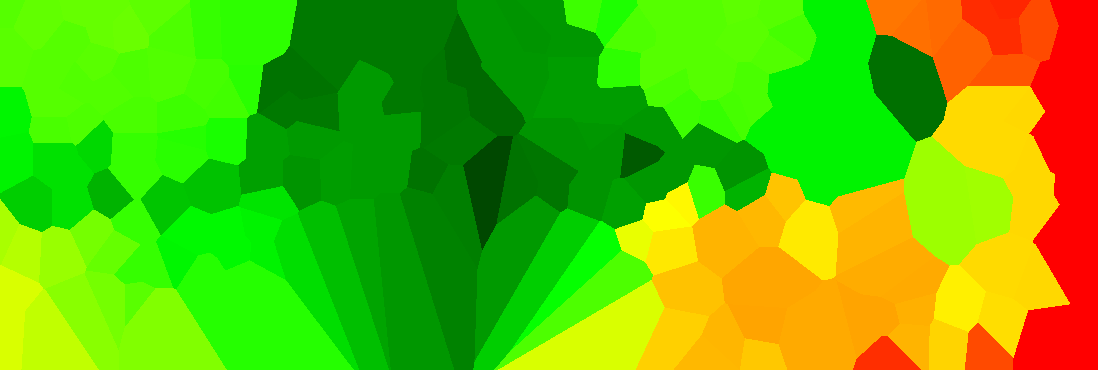} 
	\includegraphics[width=.24\linewidth]{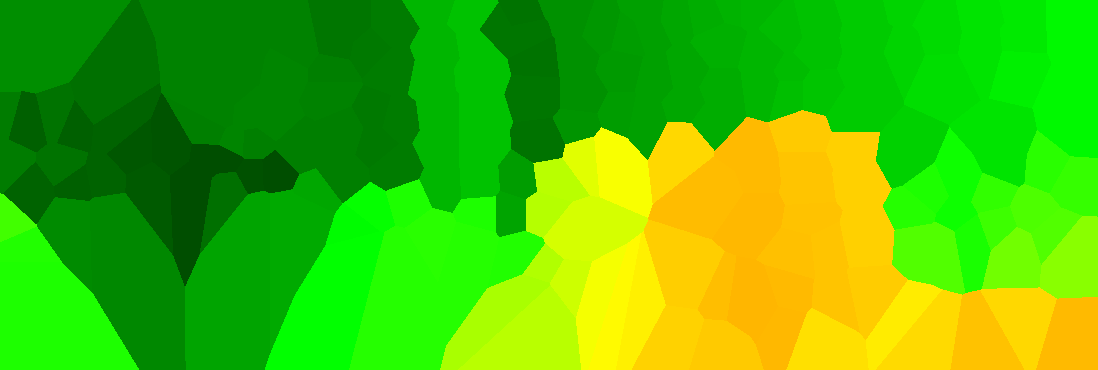}
	\includegraphics[width=.24\linewidth]{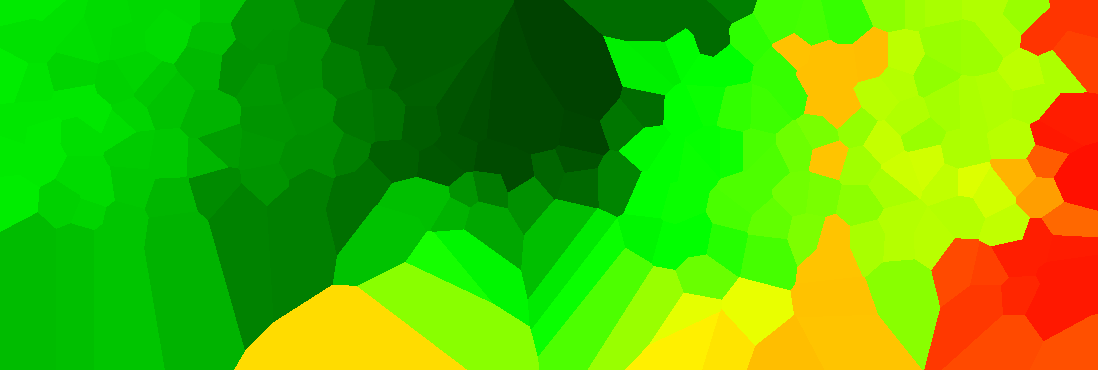}
	\includegraphics[width=.24\linewidth]{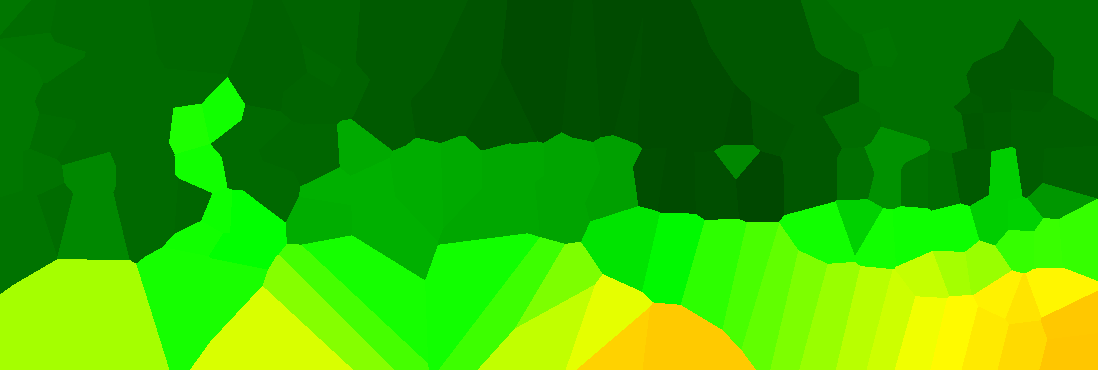} \\
	\caption{Nearest neighbor interpolation based on the same data points,
		which are used for the PCA based interpolation 
		in figure~\ref{fig:experiments:kitti} and \ref{fig:experiments:vkitti}.
		The first row correspond to the KITTI dataset examples and the second to the VKITTI ones.}
	\label{fig:experiments:nearestneighbor}
\end{figure*}

The evaluation of our method is performed on all 11 test sequences of the KITTI odometry benchmark
 \cite{GeigerLenzUrtasun2012CVPR}
 and on the 5 synthetic sequences of the Virtual KITTI dataset \cite{GaidonWangCabonVig2016},
  which provides ground truth depth information on pixel level.
We evaluate our estimated depth map in terms of 2D and 3D projection errors and only for valid points.
This means, that only points are taken into account which are in the field of view of the camera
 after projection them from frame $ A $ to another frame $ B $ and valid ground truth data exists.
During all our experiments, we set the measurement noise to $ \sigma_z = 2 $ px.

Furthermore, we compare our results with the nearest neighbor interpolation for the same sparse data points.
Finally, in a further experiment,
our approach uses the depth values of sparse discrete feature points 
from a state-of-the-art monocular SLAM algorithm \cite{FananiOchsBradlerMester2016IV}.
This demonstrates that dense interpolated depth maps can be computed with sparse measured depth values in practice.

\subsection{Learning the PCA Basis}

For the offline learning of the PCA basis,
we use the 10 training sequences of the KITTI odometry benchmark \cite{GeigerLenzUrtasun2012CVPR}.
This dataset consists of 23201 images,
which is equal to the number of our learned PCA basis vectors.
We do not use the training set for optical flow or stereo benchmark,
because these datasets consists only of few images and sparse ground truth depth from a LIDAR scanner;
this does not provide enough data and enough variation to learn the PCA basis.

Unfortunately, the ground truth depth information is not available for the odometry benchmark.
Thus, we computed the depth maps with the semi-global block matching (SGBM) algorithm
 by Hirschmueller \cite{Hirschmueller2008}.
Note that we are using  disparity values for all depth maps listed below.
Due to stereo ambiguities, stereo shadows and regions
where SGBM cannot find a match,
these depth maps typically contain a number of known invalid pixels.
Since the PCA basis should not be trained with such invalid regions,
we have interpolated these regions with a standard nearest neighbor interpolation.
Furthermore, we used a $ 5 \times 5 $ box filter to blur the depth maps,
because we cannot expect the stochastic PCA model
to represent fine structured elements in the depth field.

Finally, the PCA basis has been learned on 23201 depth maps.
Hence, the complete PCA basis consists of 23201 eigenvalues and eigenvectors.
In the following, for reasons of computational effort,
we use only the 500 largest eigenvalues and their corresponding eigenvectors.
This allows us to reconstruct approximately 95\% of the information within a depth map.

\subsection{Evaluation Measures}

The estimated depth maps are evaluated through the experiments
based on the 2D and 3D projection errors as evaluation measures.
Since we used disparities $ d $ as depth values for learning the PCA basis,
and since both the baseline $ b $ and the focal length $ f $ of the underlying stereo camera is known,
the actual depth $ w $ is derived from $ w = \frac{b \cdot f}{d} $.

Given a 2D image point $ \vec{x} $ and the known camera matrix~$\mathbf{K}$,
the corresponding 3D point $ \vec{X} $ can be computed as:
\begin{align}
	\vec{X} = w \cdot \mathbf{K}^{-1} \cdot \begin{pmatrix}
		\vec{x} \\ 1
		 \end{pmatrix}.
\end{align}
A 3D point $ \vec{X}_A $ in frame $ A $ can be projected into another frame $ B $ 
with a given projection matrix $ \mathbf{P} \in \mathbb{R}^{4 \times 4} $ using the following equation:
\begin{align}
	\begin{pmatrix}
		\vec{X}_B \\
		1
	\end{pmatrix}
	= \mathbf{P} \cdot
	\begin{pmatrix}
		\vec{X}_A \\
		1
	\end{pmatrix}.
\end{align}
The re-projection of a 3D point $ \vec{X} $ into the image coordinate system
with the camera matrix $ \mathbf{K} $ is given by the projection operation $ \pi(\cdot) $.

For an arbitrary image point in frame $ A $,
a reference 3D point $ \vec{X}_{A, ref} $, where the depth is obtained by ground truth data,
and an estimated 3D point $ \vec{X}_{A, est} $ given the interpolated depth is computed.
These two 3D points are projected into frame $ B $.
The two evaluation measures are then defined as follows:
\begin{align}
	\Delta_{3D} &= || \vec{X}_{B, ref} - \vec{X}_{B, est}||^2_2 \\
	\Delta_{2D} &= || \pi(\vec{X}_{B, ref}) - \pi(\vec{X}_{B, est})||^2_2.
\end{align}
Hence, the 3D error $ \Delta_{3D} $ measures the deviation
between the projected ground truth 3D point and the estimated one.
The 2D error $ \Delta_{2D} $ is a measure for the mismatch of a projected point 
into the image $ B $ from the ground truth position of the true position in the image.

\subsection{KITTI Dataset}
The KITTI odometry test dataset \cite{GeigerLenzUrtasun2012CVPR} 
contains 11 sequences with more than 29000 images,
which show different scenes like driving on urban,
rural, or highway roads.
This variety of scenes is challenging for our method,
because the learned PCA basis must represent all these unseen scenes.

Since KITTI does not provide the ground truth projection parameters between two consecutive temporal frames,
we cannot evaluate our approach temporally.
Therefore we used the known stereo projections and evaluate our method
only between the two cameras of the stereo system.
This does not mean that our method is limited to stereo.
It depends only on a sparse set of depth measurements.
For generating such a set, we use the \emph{'good features to track'} algorithm by \cite{ShiTomasi}.
The depth values for each detected feature point are obtained by the depth map which is computed by SGBM.
Examples of the detected feature points
and the SGBM depth maps are shown in the first and second column in figure \ref{fig:experiments:kitti}.

Even though we have a small basis of only 500 eigenvectors,
it is possible to reconstruct quite well from only a few measurements 
as long as they are distributed reasonably across the image.
Some interpolation examples are shown in the third column of figure \ref{fig:experiments:kitti}.
In the corresponding uncertainty images $ \vec{\xi} $ (last column),
it can be seen that our interpolation method is uncertain in regions
where nearly no measurements located, like in the sky.
Thus, the uncertainty can serve as a measure
how good the depth can be estimated without knowing the ground truth.
This assumption is also supported by the first histogram in figure \ref{fig:experiments:evaluation}.
The 2D error $ \Delta_{2D} $ is relative small in the first bins and increases with the uncertainty intervals.
However, more than 60\% of all evaluated points from the datasets
fall into the first two bins, which indicates that the uncertainty images as well as the interpolated depth maps perform well.

The other two histograms in the first row of figure \ref{fig:experiments:evaluation} reveals
that the errors of $ \Delta_{2D} $ and $ \Delta_{3D} $ are small, when the points are close to the camera. 
Not surprisingly, if the points are far away, the errors usually increase.

\subsection{Virtual KITTI Dataset}

We have evaluated our approach on all five public available sequences of the Virtual KITTI dataset
\cite{GaidonWangCabonVig2016} as well,
which comprises driving scenes on urban and rural roads. 
We take all 2359 synthetic generated images and use 'overcast' as weather category.
The learned PCA basis is the same like in the previous KITTI experiment.
Thus, we have not learned a specific basis for VKITTI, which also shows the ability of generalization. 
However, the quantitative and qualitative results remained the same or are even better. 
The reason for this is probably the exact ground truth at pixel level. 
In contrast to the KITTI dataset, where we used the computed SGBM depth map as ground truth,
VKITTI provides exact depth maps and poses.
Another advantages is that we do not get wrong depth measurements for the interpolation.
This could happen at the KITTI dataset,
if a pixel location was selected where SGBM returned a false depth value.
This may lead to a possibly bad representation in terms of
the basis coefficients.

Due to presence of ground truth pose files and the non existence of stereo images in VKITTI,
we evaluate our method in temporal direction (mono case).
This implies that we use the available ground truth pose to project the feature points
from frame $ A $ at time $ t $ to frame $ B $ at time $ t+1 $.
We use the same GFTT feature detector
as in the KITTI dataset.
In figure \ref{fig:experiments:vkitti},
some exemplary results are shown including the ground truth depth map of VKITTI
and the uncertainty image of our interpolated depth map.

For all images of the VKITTI dataset, the 2D error $ \Delta_{2D} $ of most image points
falls into the first error interval,
which means that $ \Delta_{2D} < 1 \textnormal{px}$.
This is independent of the uncertainty or depth intervals,
as it is shown by the second row of figure \ref{fig:experiments:evaluation}. 
So, the estimated depth values are very close to the ground truth data on nearly all possible points.
There are hardly any outliers.
Additionally, if only the first two uncertainty intervals are used, more than 50\% of the points are captured.
Roughly 80\% of these points are estimated with a deviation to the ground truth with less than 1 pixel.

The 3D error $ \Delta_{3D} $ is small for points which are close to the camera.
But there are also a large number of points in the far-field,
where the error is quite large.
This can be explained by the ground truth depth maps from VKITTI.
They have limited the maximum distance to the camera at $655.35$ meters.
This is not equal to the KITTI dataset, where the maximum distance is greater.
Thus, the far-field points can be estimated with our learned PCA basis further away
than it is encoded in the VKITTI depth maps,
which leads to a significant error for points in the far field.

\subsection{Comparison to Nearest Neighbor Interpolation}

\begin{figure}[htbp]
	\centering
	\includegraphics[width=.9\linewidth]{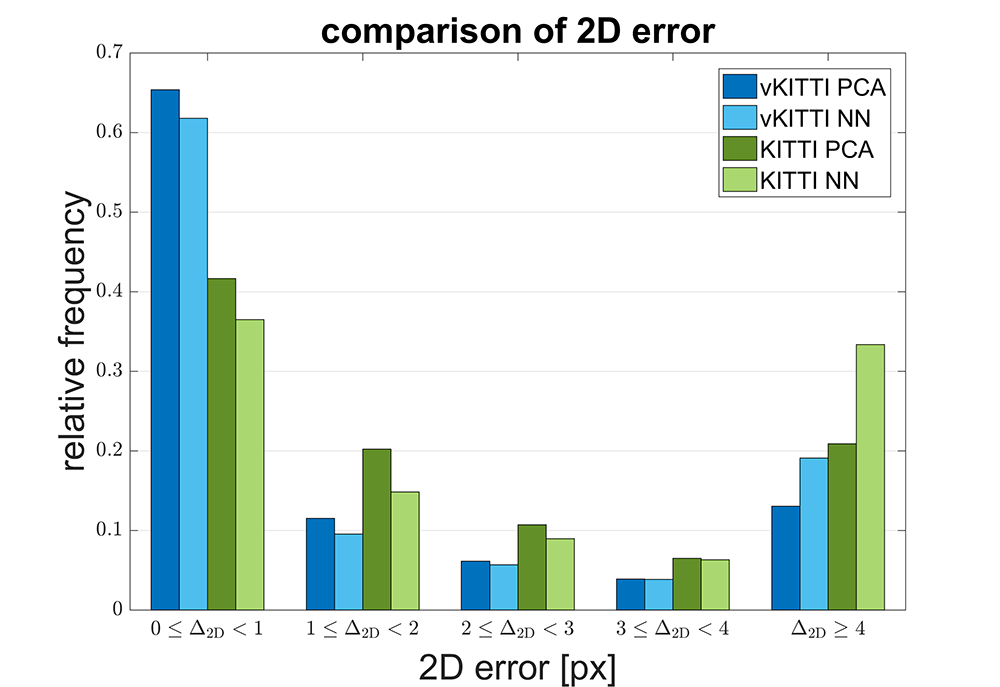}
	\caption{Distribution of the 2D error $ \Delta_{2D} $ 
	for our approach and the nearst neighbor interpolation throughout both evaluated dataset.}
	\label{fig:experiments:nnausweiterung}
\end{figure}

In this section, we compare our method to the nearest neighbor interpolation.
For both methods, we use the same data points, of course.
Figure \ref{fig:experiments:nearestneighbor} shows the nearest neighbor interpolation for the data,
which we used for the examples in the KITTI and VKITTI experiments.
In comparison to our method, the nearest neighbor interpolation is much coarser.
Thus, the depth impression is not as good as in our method.
This also exhibits the evaluation of the 2D error $ \Delta_{2D} $ throughout both datasets.
In the first two bins of the histogram \ref{fig:experiments:nnausweiterung}, where $ \Delta_{2D} < 2  $,
more points of our methods than from the nearest neighbor interpolation satisfy this condition.
Additionally, in the last interval, where the 2D error is greater than 4 pixels,
more points of the nearest neighbor interpolation are accumulated than from our approach.

Moreover, the mean of the 2D error $ \Delta_{2D} $ for the nearest neighbor interpolation within both datasets
 is nearly 50\% as larger compared to our interpolation.
For the KITTI dataset, the mean value for the PCA method is $ 3.4 $ px and for the nearest neighbor $ 5.4 $ px.
Similarly, the mean value for the VKITTI dataset is $ 2.0 $ px for our approach and $ 3.3 $ px for the other one.

\subsection{PCA Interpolation in Practice}

During all experiments so far, we used as sparse measurements always ground truth depth values.
For the KITTI dataset, the SGBM depth values acted as ground truth values.
However, these depth measurements are not correct in all cases either.
Nevertheless, our method should also work with sparse data from a state-of-the-art monocular SLAM method.

\begin{figure}[ht]
	\centering
	\includegraphics[width=.49\linewidth]{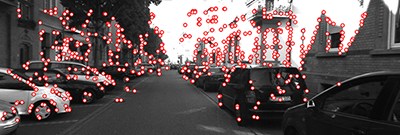}
	\includegraphics[width=.49\linewidth]{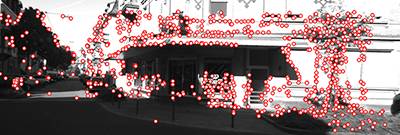} \\
	\vspace{0.1cm}
	\includegraphics[width=.49\linewidth]{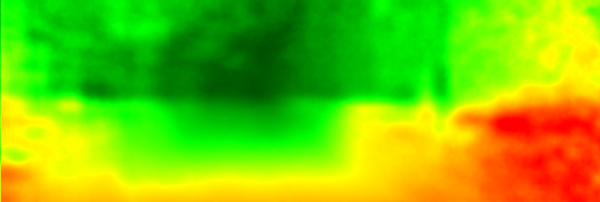}
	\includegraphics[width=.49\linewidth]{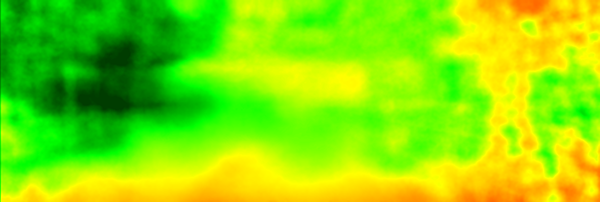}
	\caption{Exemplary results of the dense PCA interpolation with sparse measurements
	using the monocular SLAM algorithm from \cite{FananiOchsBradlerMester2016IV}.}
	\label{fig:experiments:pbt}
\end{figure}

Two exemplary results of a dense PCA interpolation with the sparse depth values
from the propagation based tracking (PbT) algorithm \cite{FananiOchsBradlerMester2016IV} as input measurements
 are shown in figure \ref{fig:experiments:pbt}. 
These examples come from the sequence 13 of the KITTI odometry test dataset \cite{GeigerLenzUrtasun2012CVPR}.
The PbT method measures in these two frames roughly 400 feature points,
which are represented in the first row of figure \ref{fig:experiments:pbt}.
Based on these sparse measurements, 
our PCA interpolation computes a dense depth map,
which is shown in the second column of the same figure.
These examples demonstrate that our interpolation is not limited to depth measurements which come from ground truth data.
The depth impression of the interpolated depth map is coherent for non ground truth depths values, too.
Hence, these computed dense depth maps can be used in practice for initialization purposes for dense matching or tracking approaches,
like also claimed in \cite{WulffBlack2015}.
\section{Summary \& Conclusion}

In this work, we introduced a novel method to estimate dense depth maps from highly sparse measurements.
The proposed interpolation uses a statistical model to obtain
reduced rank signal
subspaces which are learned by PCA.
The dense depth maps are reconstructed as weighted linear combinations of these PCA basis signals.
The necessary coefficients are determined by a maximum a-posteriori estimation approach which defuses the otherwise underdetermined problem.

The resulting depth maps yield a convincing coarse impression of the underlying depth structure.
The numerical evaluation also shows that the estimated dense depth maps
approximate the depth well on two challenging automotive datasets.
Furthermore, we introduce uncertainty maps, a self-diagnosis tool
that allows to find those areas
where the interpolated / extrapolated depth is
questionable.
A drawback of these uncertainty maps are their relatively large computation costs.
If the computation of the uncertainty map is skipped, our depth map interpolation scheme runs in real-time on a standard desktop computer.

A typical application case of our method 
is to provide a first initialization for a further densification step.
For instance, a dense optical flow algorithm \cite{Horn81Schunck1981}
converges faster utilizing a good initialization.
This process flow is also claimed in \cite{WulffBlack2015}.
Furthermore, sparse tracking algorithms \cite{FananiOchsBradlerMester2016IV,EngelSchoepsCremers2014,BradlerOchsMester2017WACV}
can be initialized with our dense depth
in order to give good initializations
for tracking further feature points.
These application perspectives show that
the proposed dense depth may be a valuable new component
for real-time dense reconstruction of the environment in context of autonomous driving.




\end{document}